\documentclass[letterpaper, 10 pt, conference]{ieeeconf}  

\IEEEoverridecommandlockouts                              

\usepackage{graphicx} 
\usepackage{amsmath} 

\usepackage{amsthm}
\usepackage{listings}
\theoremstyle{definition}

\usepackage{amssymb}
\usepackage{pifont}
\usepackage{booktabs}
\usepackage{csvsimple}

\usepackage{caption}
\usepackage{subcaption}
\let\labelindent\relax
\usepackage{enumitem}

\usepackage{algorithm}
\usepackage[noend]{algpseudocode}
\algnewcommand{\LineComment}[1]{\State \(\triangleright\) #1}
\usepackage[colorlinks=true]{hyperref}
\usepackage[font=scriptsize,labelfont=bf]{caption}
\usepackage[dvipsnames]{xcolor}

\usepackage{etoolbox}
\makeatletter
\patchcmd{\@makecaption}
  {\scshape}
  {}
  {}
  {}
\makeatother

\title{\LARGE \bf
Why did I fail? A Causal-based Method to Find Explanations for Robot Failures
}

\author{Maximilian Diehl and Karinne Ramirez-Amaro 
\thanks{Maximilian Diehl and Karinne Ramirez-Amaro. Faculty of Electrical Engineering, Chalmers University of Technology, SE-412 96 Gothenburg, Sweden.
        {\tt\small \{diehlm, karinne\}@chalmers.se}}%
}

\begin{document}

\maketitle
\thispagestyle{empty}
\pagestyle{empty}

\begin{abstract} 

Robot failures in human-centered environments are inevitable. Therefore, the ability of robots to explain such failures is paramount for interacting with humans to increase trust and transparency. To achieve this skill, the main challenges addressed in this paper are I) acquiring enough data to learn a cause-effect model of the environment and II) generating causal explanations based on that model. We address I) by learning a causal Bayesian network from simulation data. Concerning II), we propose a novel method that enables robots to generate contrastive explanations upon task failures. The explanation is based on setting the failure state in contrast with the closest state that would have allowed for a successful execution. This state is found through breadth-first search and is based on success predictions from the learned causal model. We assessed our method in two different scenarios I) stacking cubes and II) dropping spheres into a container. The obtained causal models reach a sim2real accuracy of 70\% and 72\%, respectively\footnote{A visual demonstration of our method and the real-world experiments can be found under \url{https://youtu.be/rl_zDfUZ2Kk}}. We finally show that our novel method scales over multiple tasks and allows real robots to give failure explanations like 'the upper cube was stacked too high and too far to the right of the lower cube.' 

\end{abstract}


\section{Introduction}
One important component in human interactions is the ability to explain one's actions, especially when failures occur~\cite{Miller19Explanation, Hellstroem21}. It is argued that robots need this skill if they were to act in human-centered environments on a daily basis~\cite{bookOfWhy}. Moreover, explainability is shown to increase trust and transparency in robots~\cite{Miller19Explanation, Hellstroem21}, and the diagnoses capabilities of a robot are crucial for correcting its behavior~\cite{Mitrevski20IROS}.

\begin{figure}[t!]
\centering
  \includegraphics[width=0.48\textwidth]{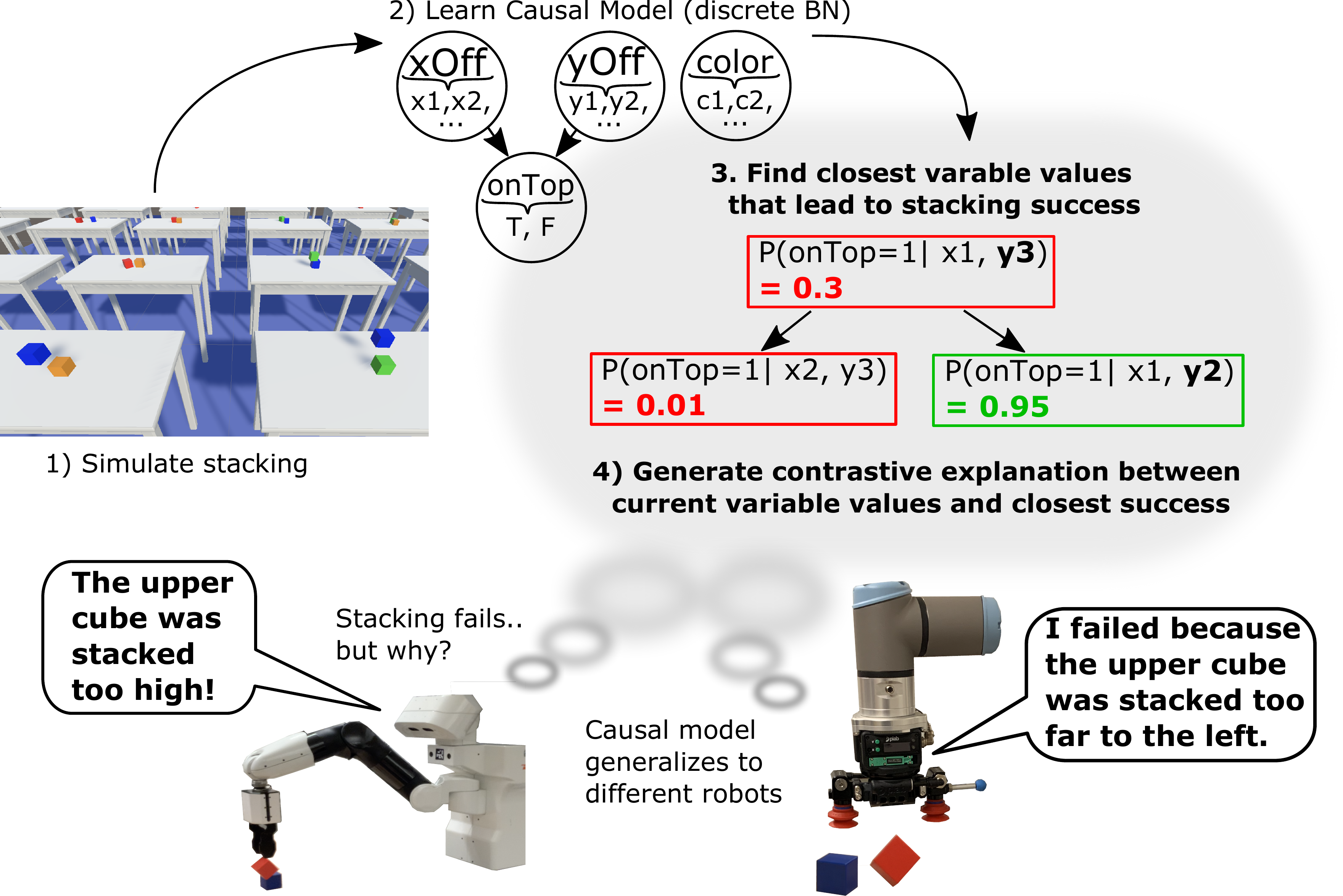}
  \caption{Depicts our method that allows robots to explain their failures. First, we learn a causal model from simulations (steps 1,2). A contrastive explanation is generated upon task failures (steps 3,4). Finally, the obtained models are evaluated on two different tasks (cube stacking and sphere dropping) and transferred to two different robots that provide explanations when they commit errors.}
\label{fig:overview}
\vspace{-5mm}
\end{figure}

There are different types of failures, e.g., task recognition errors (an incorrect action is learned) and task execution errors (the robot drops an object) \cite{LeeDemiris2013, karapinar2015}. In this work, we focus on explaining execution failures. For example, a robot is asked to stack two cubes (see Fig. \ref{fig:overview}). The robot will first pick
up a cube and move its gripper above the goal cube. However, due to sensor and motor inaccuracies, the robot places its gripper slightly shifted to the left, which results in an imperfect cube alignment between the cubes. 
Therefore, 
the upper cube lands on the goal but bounces to the table. In such a situation, we expect the robot to reason about what went wrong and generate an explanation based on its previous experience, e.g., 'I failed because the upper cube was dropped too far to the left of the lower cube.'

Typically, explanations are based on the concept of causality~\cite{LewisCausalExplanation}. Obtaining a causal model of the environment is often addressed through statistical methods that learn a mapping between possible causes (preconditions) and the action-outcome~\cite{Mitrevski20IROS, Bauer20ICRA}. However, such statistical models alone are not explanations in itself~\cite{Miller19Explanation} and require another layer that interprets these models to produce explanations. Another problem is that a considerable amount of data is needed to learn cause-effect relationships. In this case, training such models using a simulated environment will allow a faster and more extensive experience acquisition~\cite{Mitrevski20IROS}.

In this paper, we propose a method for generating causal explanations of failures based on a causal model that provides robots with a partial understanding of their environment (see Fig.~\ref{fig:overview}). First, we learn a causal Bayesian network from simulated task executions, tackling the problem of knowledge acquisition. We also show that the obtained model can transfer the acquired knowledge (experience) from simulation to reality and is agnostic to several real robots with different embodiments. Second, we propose a new method to generate explanations of execution failures based on the learned causal knowledge. Our method is based on a contrastive explanation comparing the variable parametrization associated with the failed action with its closest parametrization that would have led to a successful execution, which is found through breadth-first search (BFS). Finally, we analyze the benefits of this method in two different scenarios: I) stacking cubes and II) dropping spheres into a container.

To summarize, our contributions are as follows:
\begin{itemize}
\item We present a novel method to generate contrastive causal explanations of action failures based on causal Bayesian networks. 
\item We demonstrate how causal Bayesian networks can be learned from simulations, exemplified in a cube stacking and sphere dropping scenario and provide extensive real-world experiments that show that the obtained causal models are transferable from simulation to reality without any retraining. Our method is agnostic to various robot platforms with different embodiments and scales over multiple tasks and scenarios. We, thus, show that the simulation-based model serves as an excellent prior experience for the explanations, making them more generally applicable.
\end{itemize}

\section{Related Work}
\subsection{Causality in Robotics}
Despite being acknowledged as an important concept, causality is relatively underexplored in the robotics domain~\cite{Brawer20IROS, Hellstroem21}. Some works explore causality to distinguish between task-relevant and -irrelevant variables~\cite{Stocking22}. For example, CREST~\cite{crest} uses causal interventions on environment variables to discover which of the variables affect an RL policy. They find that excluding them impacts generalizability and sim-to-real transfer positively. In~\cite{Bhat2016HumanoidIA} a set of causal rules is defined to learn to distinguish between unimportant features in physical relations and object affordances. 
Brawer et al. present a causal approach to tool affordance learning~\cite{Brawer20IROS}. Some works explore Bayesian networks, for example, to learn statistical dependencies between object attributes, grasp actions, and a set of task constraints from simulated data~\cite{Song10}. While the main objective is to use graphical models to generalize task executions, these works don't look into the question of how these models can be utilized for failure explanations.
A different paper~\cite{Uhde20RobotScientist} investigates the problem of learning causal relations between actions in household-related tasks. They discover, for example, that there is a causal connection between opening a drawer and retrieving plates from human demonstrations.
They only retrieve causal links between actions, while we focus on causal relations between different environment variables, like object features and the action outcome.

\subsection{Learning explainable models of cause-effect relations}

In the planning domain, cause-effect relationships are represented through (probabilistic) planning operators~\cite{Diehl21}.
Mitrevksi et al. propose the concept of learning task execution models, which consists of learning symbolic preconditions of a task and a function approximation for the success model~\cite{Mitrevski20IROS}, based on Gaussian Process models. They noted that a simulated environment could be incorporated for a faster and more extensive experience acquisition, as proposed in~\cite{Song10}. Human virtual demonstrations have been used to construct planning operators to learn cause-effect relationships between actions and observed state-variable changes~\cite{Diehl21}. However, even though symbolic planning operators are considered human-understandable, they are not explanations in itself, thus requiring an additional layer that interprets the models and generates failure explanations.

Some other works also aim to learn probabilistic action representations experience to generalize the acquired knowledge. For example, learning probabilistic action effects of dropping objects into different containers \cite{Bauer20ICRA}. Again, the main objective is to find an intelligent way of generalizing the probability predictions for a variety of objects, e.g., bowl vs. bread box, but their method does not include any understanding of why there is a difference in the dropping success probabilities between these different objects. 

\subsection{Contrastive Explanations}
Contrastive explanations are deeply rooted in the human way of generating explanations~\cite{Miller19Explanation}. This also had a significant impact on explanation generation in other fields like Explainable AI Planning (XAIP)~\cite{XAIP}. In XAIP, typical questions that a machine should answer are \textit{why a certain plan was generated vs. another one?} or \textit{why the plan contains a particular action $a_1$ and not action $a_2$?}~\cite{XAIP, Seegebarth12}. We, however, are mostly interested in explaining why specific actions failed based on environment variables like object features. 
A method for explaining synthesis failures of high-level robot task specifications (encoded through LTL formulae) is presented in~\cite{Vasumathi12}. However, the causes need to be explicitly modeled (violations of specification constraints), while, in our approach, the causes are automatically detected during the BN learning process. Das et al. generate verbal failure explanations~\cite{Das21}, by learning an encoder-decoder network that maps current information about the robot and environment state into a vector of words. 
However, the method does not scale well since it requires data with annotations about each failure cause. Our approach only requires annotations regarding the action success, which can be binary and are generally easier obtainable. Additionally, we encode the explanations directly in the causal structure of the different state variables instead of learning a black-box model. In a follow-up study~\cite{Das21b}, the authors use MOTIFNET~\cite{Zellers18} to autonomously detect spatial relationships and object attributes in a given scene. Then, pairwise ranking is used to filter out the subset of relevant relations for a particular explanation. Annotations for pairwise preferences of one relation over another need to be provided for training an SVM, which cannot be easily automated since they require human input. Due to the close relation to our approach, we discuss these works~\cite{Das21, Das21b} in more detail in Sec.~\ref{sec:baseline}.

\section{Our approach to explaining failures}
Our proposed approach consists of three main steps: A) Identification of the variables used in the analyzed task; B) Learning a Bayesian network which requires to 1) Learn a graphical representation of the variable relations (structure learning) and 2) to learn conditional probability distributions (parameter learning); and C) Our proposed method to explain failures, based on the previously obtained model.

\subsection{Variable definitions and assumptions}
Explaining failures, requires to learn the connections between possible causes and effects of an action. We describe an action via a set of random variables $\mathrm{\textbf{X}} = \{ X_1, X_2,..., X_n \}$, which need to be defined by the experiment designer during the experiment setup. We require $\mathrm{\textbf{X}}$ to contain a set of treatment variables $C \subset \mathrm{\textbf{X}}$, which describe potential causes, and outcome (effect) variables $E \subset \mathrm{\textbf{X}}$. Then, the goal of causal inference is to estimate the effect of $C$ on $E$~\cite{Vowels22}.

Data samples for learning the causal model can, in principle, be collected in simulation or the real world. A data sample $d$ consists  of a particular parametrization of 
$\mathrm{\textbf{X}}$, which we define as $d = \{ X_1 = x_{1}, X_2 = x_{2}, ..., X_n = x_{n} \}$, where $n$ denotes the number of variables. It is important to sample values for possible causes $C$ randomly. Randomized controlled trials are referred to as the gold standard for causal inference~\cite{bookOfWhy} and allow us to exclude the possibility of unmeasured confounders. Consequently, all detected relations between the variables $\mathrm{\textbf{X}}$ are indeed causal and not merely correlations. Besides the apparent advantage of generating truly causal explanations and avoiding the danger of possible confounders, causal models can also answer interventional questions. In contrast, non-causal models can only answer observational queries. The experiment must satisfy the sampled variable values before executing the action for data collection. $E$ is measured at the end of the experiment.

We define another set $X_{goal} = \{$ $d_{goal_1},$  $d_{goal_2},$ $...,d_{goal_h}\}$ that contains all possible variable parametrizations that denote a successful action execution. Then, an action is successful \textit{iff} its parametrization $d \in X_{goal}$. Note, that it is out of scope of this paper, to discuss methods that learn $X_{\text{goal}}$, but rather assume $X_{\text{goal}}$ to be provided a priori. In other words, we assume that the robot knows how an unsuccessful task execution is defined in terms of its outcome variables $E$ and is thus able to detect it by comparing the action execution outcome with $X_{\text{goal}}$. Note, however, that the robot has no a-priori knowledge about which variables in $\mathrm{\textbf{X}} = {X_1, X_2, ..., X_n}$ are in $C$ or $E$, nor how they are related. This knowledge is generated by learning the Bayesian network.

To efficiently learn a Bayesian network, some assumptions are needed to handle continuous data~\cite{ChenDisc17}, mainly because many structure learning algorithms do not accept continuous variables as parents of discrete/categorical variables~\cite{bnlearn}. In our case, this means that some effect variables from $E$ could not have continuous parent variables out of $C$, which would likely result in an incorrect Bayesian network structure. As a preprocessing step, we therefore discretize all continuous random variables out of $\mathrm{\textbf{X}}$ into intervals with an equal number of samples.

\subsection{Our proposed pipeline to learn the causal model}
Formally, Bayesian networks are defined via a graphical structure $\mathcal{G} = (\mathrm{\textbf{V}}, A)$, which is a \textit{directed acyclic graph} (DAG), where $\mathrm{\textbf{V}} = \{ X_1, X_2, ..., X_n \}$ represents the set of nodes and $A$ is the set of arcs~\cite{bnlearn}. Each node $X_i \subseteq\  \mathrm{\textbf{X}}$ represents a random variable. Based on the dependency structure of the DAG and the \textit{Markov property}, the \textit{joint probability distribution} of a Bayesian network can be factorized into a set of \textit{local probability distributions}, where each random variable $X_i$ only depends on its direct parents $\Pi_{X_i}$:
\begin{equation}\label{eq:bn} 
    P(X_1, X_2, ..., X_n) = \prod_{i=1}^{n}P(X_i|\Pi_{X_i})
\end{equation}
Learning a Bayesian network from data consists of two steps:  
\subsubsection{Structure Learning}
\label{sec:structureLearning}
The purpose of this step is to learn the graphical representation of the network $\mathcal{G} = (\mathrm{\textbf{V}}, A)$ and can be achieved by a variety of different algorithms. An extensive survey of potentially equally valid structure learning algorithms, like~\cite{Xun18} is presented in~\cite{Vowels22}. For the remainder of this paper, we choose the Grow-Shrink~\cite{gs} algorithm (gs) to learn $\mathcal{G}$. gs falls into the category of \textit{constraint-based-algorithms}, which use statistical tests to learn conditional independence relations (also called "constraints") from the data~\cite{bnR}. Note that learning plausible assumptions about causal relations is one of the biggest challenges in the whole process of causal inference~\cite{Sharma21}. For example, in some cases it is challenging to determine the direction of causal relations purely from the joint distribution of the observational data (thus without additional interventional experiments, additional domain knowledge, or certain assumptions about the data distribution)~\cite{Peters17}. Structure learning is an active field of research~\cite{Sharma21}, and
this paper will use the learned structure to generate causal-based explanations of failures.
Therefore, we assume that the outcome of the structure learning step is indeed the correct Bayesian network graph $\mathcal{G}$, or has been manually revised based on domain knowledge.


\subsubsection{Parameter Learning}
The purpose of this step is to fit functions that reflect the \textit{local probability distributions}, of the factorization in formula (\ref{eq:bn}). We utilize the maximum likelihood estimator for conditional probabilities (\textit{mle}) to generate a conditional probability table based on the previously obtained network structure.


\subsection{Our proposed method to explain failures}
\begin{figure}[ht!]
\centering
  \includegraphics[width=0.48\textwidth]{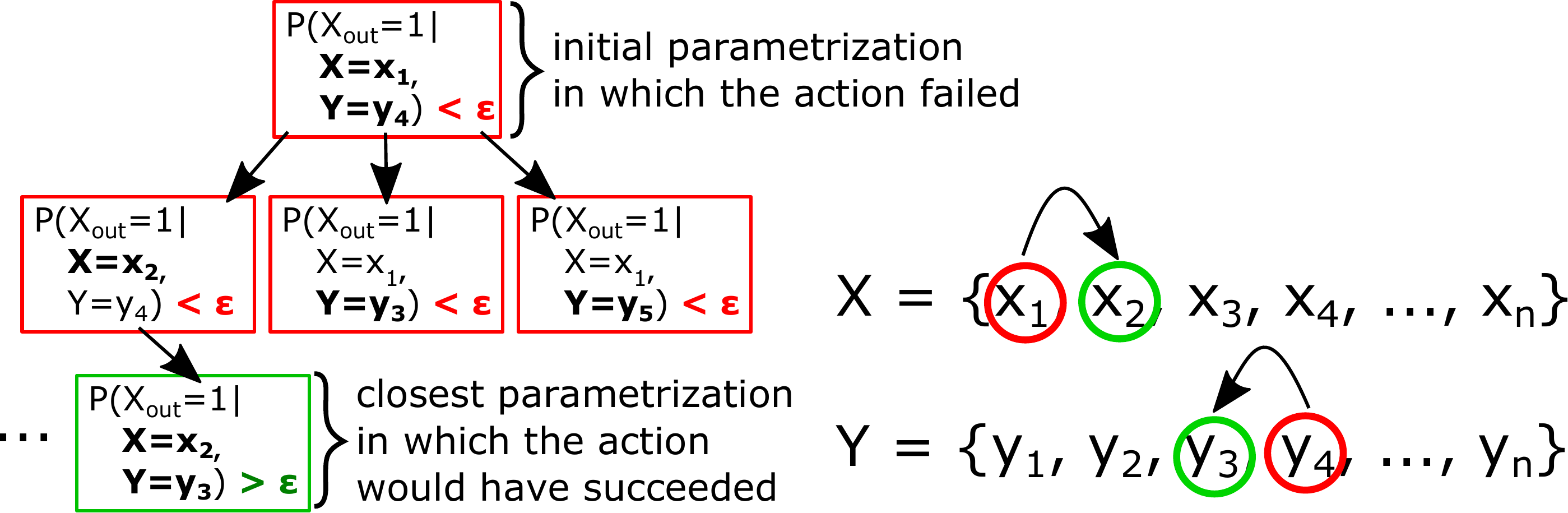}
  \caption{Exemplifies how contrastive explanations are generated from the BFS search tree.}
\label{fig:explainGen}
\vspace{-5mm}
\end{figure}

Our proposed method to generate contrastive failure explanations uses the obtained causal Bayesian network to compute success predictions and is summarized in algorithm~\ref{alg:explanation}.
\setlength{\textfloatsep}{2pt}
\begin{algorithm}
  \caption{\label{alg:explanation}Failure Explanation}
  \hspace*{\algorithmicindent} \textbf{Input:} failure variable parameterization $x_{\text{failure}}$, graphical model $\mathcal{G}$, structural equations $P(X_i|\Pi_{X_i})$, discretization intervals of all model variables $X_{\text{int}}$, success threshold $\epsilon$, goal parametrizations $X_{goal}$ \\
  \hspace*{\algorithmicindent} \textbf{Output:} solution variable parameterization $x_{\text{solution}_{\text{int}}}$, solution success probability prediction $p_{\text{solution}}$
  \begin{algorithmic}[1]
  \State $x_{\text{current}_{\text{int}}} \leftarrow \textsc{getIntervalFromValues}(x_{\text{failure}}, X_{\text{int}})$
  \State $P \leftarrow \textsc{generateTransitionMatrix}(X_{\text{int}})$ 
  \State $q \leftarrow [x_{\text{current}_{\text{int}}]}$
  \State $v \leftarrow []$
  \While{$q \neq \emptyset$} 
   \State $node \leftarrow \textsc{Pop}(q)$ 
   \State $v \leftarrow \textsc{Append}(v, node)$
   \ForAll{$\text{transition } t \in P(node)$} 
   \State $child \leftarrow \textsc{Child}(P, node)$
   \If{$child \not\in q,v $}
   \State $p_{\text{solution}} = P(d \in X_{goal}|\Pi_{child})$
   \If{$p_{\text{solution}} > \epsilon$}
   \State $x_{\text{solution}_{\text{int}}} \leftarrow child$
   \State $\textsc{return}(p_{\text{solution}}, x_{\text{solution}_{\text{int}}})$
   \EndIf
   \State$ q \leftarrow \textsc{Append}(q, x_{\text{current}_{\text{int}}})$
   \EndIf
   \EndFor
   \EndWhile
  \end{algorithmic}
\end{algorithm}
In (L-2 Alg. \ref{alg:explanation})), a matrix is generated which defines transitions for every single-variable change for all possible variable parametrizations. For example, if we had two variables $X_1, X_2$ with two intervals $x', x''$. Then, the possible valid transitions for $node = (X_1=x', X_2=x')$ would be $child_1 = (X_1=x', X_2=x'')$ or $child_2 = (X_1=x'', X_2=x')$. Lines 5-15 (Alg. \ref{alg:explanation}) describe the adapted BFS procedure, which searches for the closest variable parametrization that fulfills the goal criteria of $P(d \in X_{goal}|\Pi_{child}) > \epsilon$, where $\epsilon$ is the success threshold, which can be heuristically set. The concept of our proposed method is to generate contrastive explanations that compare the current variable parametrization associated with the execution failure  $x_{\text{current}_{\text{int}}}$ with the closest parametrization that would have allowed for a succesfull task execution $x_{\text{solution}_{\text{int}}}$. Consider Figure~\ref{fig:explainGen} for a visualization of the explanation generation, exemplified on two variables $X$ and $Y$, which are both causally influencing the variable $X_{\text{out}}$. Furthermore, it is known that $x_{\text{out}} = 1 \in X_{\text{goal}}$. The resulting explanation would be that the task failed because $X=x_1$ instead of $X=x_2$ and $Y = y_4$ instead of $Y = y_3$.

\section{Experiments}
We evaluate our method to find causal explanations of failures based on two different scenarios. The goal of \textit{experiment 1} is to stack one cube on top of another. The goal of \textit{experiment 2} is to drop a sphere into different containers.


\subsection{Experiment 1: Stacking Cubes}
\begin{figure}[ht!]
\vspace{-3mm}
\centering
  \includegraphics[width=0.4\textwidth]{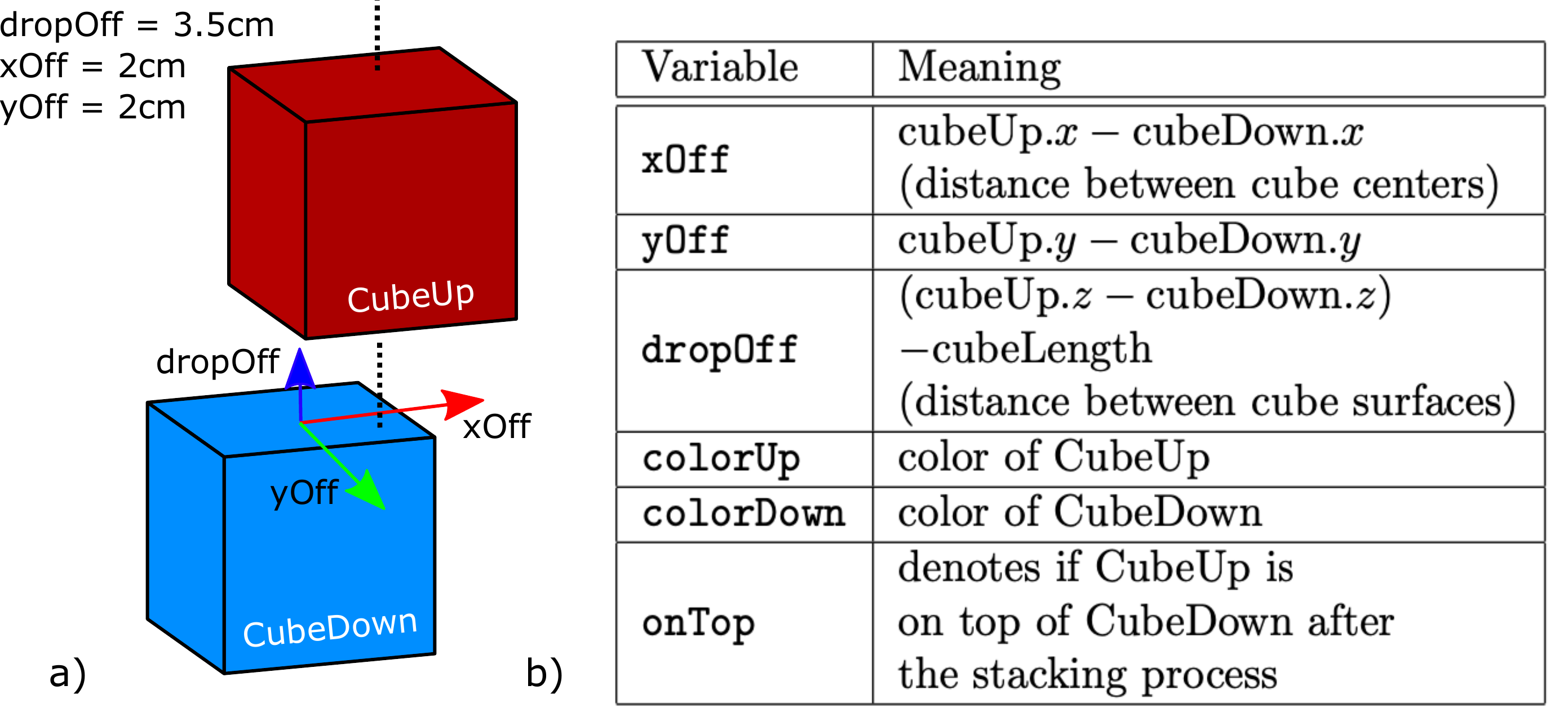}
  \caption{a) visualizes the used variables $\mathrm{\textbf{X}}$ in experiment 1 and b) describes their meaning.}
\label{fig:environment}
\end{figure}

In the cube stacking scenario, the environment contains two cubes: \textit{CubeUp} and \textit{CubeDown} (see Fig. \ref{fig:environment}). The goal of the stacking action is to place CubeUp \textit{on top} of CubeDown. We define six variables as follows: $\mathrm{\textbf{X}} = \{\texttt{xOff}, \texttt{yOff}, \texttt{dropOff}, \texttt{colorDown}, \texttt{colorUp}, \texttt{onTop}\}$. Both cubes have an edge length of 5cm.  

\subsubsection{Cube Stacking simulation setup}
We run the simulations in Unity3d which bases its physics behavior on the Nvidia PhysX engine. For training the Bayesian network we generate 20,000 samples, on 500 parallel table environments (see Fig. \ref{fig:overview}). We randomly sample values for $\texttt{xOff, yOff} \sim \mathcal{U}_{[-3, 3]}$ (in cm), $\texttt{dropOff} \sim \mathcal{U}_{[0.4, 10]}$ (in cm), $\texttt{colorUp, colorDown} = \{ \text{Red},\text{Blue}, \text{Green}, \text{Orange} \}$. $\texttt{onTop} = \{ \text{True}, \text{False} \}$ is not sampled but automatically determined after the stacking process.

\subsubsection{Cube Stacking robot experiments setup}
\label{sec:cubeStackingRobotExperiment}
We run and assess our experiments on two different robotic platforms (Fig. \ref{fig:overview}): the TIAGo service robot with one arm and parallel gripper and the UR3 with a piCOBOT suction gripper. The real cubes are 3D printed out of PLA (polylactic acid) and weigh around 25 grams each. For each robot, we run 180 stacking trials. Instead of randomly sampling values for the variables, as we do for training the causal model, we evaluate the real-world behavior at 36 different points, where $\texttt{xOff}, \texttt{yOff} = \{0, 1, 2\}$ (in cm), $\texttt{dropOff} = \{ 0.5, 2, 3.5, 5\}$ (in cm).
These 36 points were empirically chosen because they cover an area where the ideal conditions of the simulation (e.g., collisions without any rotation due to the gripper motors) could have potentially the most significant effect on behavior discrepancies. Once the upper cube is too far outside ($> 2.5$ cm), it doesn't play a role how it was dropped, so we have not included points with larger offsets than 2cm. For each unique stacking setup instantiation, we conduct five iterations. After each trial, the cubes are re-adjusted into an always similar pre-stack position by the operator. The stacking outcome (\texttt{onTop} value) was also determined by the operator. Note that the purpose of the robot experiments is not to modify the causal model that we learned from the simulation but to evaluate the model transferability to the real environment.

\subsection{Experiment 2: Dropping spheres into containers}
\begin{figure}[ht!]
\centering
  \includegraphics[width=0.44\textwidth]{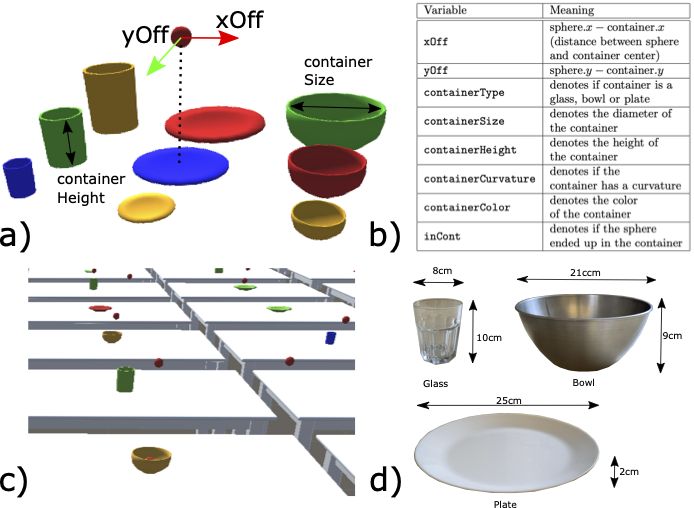}
  \caption{a) visualizes the used variables $\mathrm{\textbf{X}}$ in experiment 2, b) describes their meaning, c) visualizes the simulation setup, and d) the containers that were used for the real-world experiment.}
\label{fig:environment2}
\end{figure}

In our second experiment, the robot needs to drop spheres into different containers. The environment contains a \textit{Sphere} and one of several possible \textit{Containers}, which are shaped like a plate, bowl or glass (see Fig.~\ref{fig:environment2}). 
We define eight variables as follows: $\mathrm{\textbf{X}} = \{\texttt{xOff},$ $\texttt{yOff}, \texttt{inCont},$ $\texttt{contHeight},$ $\texttt{contSize},$ $\texttt{contType},$ $\texttt{contCurvature},$ $\texttt{contColor} \}$. This new list of variables is required because experiment 2 contains a different set of objects and new randomization parameters like the size, which we did not consider in experiment 1. The sphere has a diameter of 6.6cm and the size of the containers is randomized. We chose a constant dropping height of 0.4m.

\subsubsection{Sphere Dropping simulation setup}
For training the Bayesian network, we generate 800,000 samples, on 400 parallel table environments (see Fig.~\ref{fig:environment2}.c), in Unity3d. We randomly sample values for $\texttt{xOff, yOff} \sim \mathcal{U}_{[-6, 6]}$ (in cm), $\texttt{contColor} =$ $\{ \text{Red},$ $\text{Blue},$ $\text{Green},$ $\text{Orange} \}$, $\texttt{contType} =$ $\{ \text{Glass},$ $\text{Plate},$ $\text{Bowl} \}$. We do not directly set $\texttt{contHeight}$ and $\texttt{contSize}$, but manipulate these variables via a $\texttt{scalingFactor} \sim \mathcal{U}_{[0.4, 1]}$. In Unity3d, the scaling factor can be utilized to manipulate the size of objects. We use the same scaling factor in all three dimensions, thus the height ($\texttt{contHeight}$) to diameter ($\texttt{contSize}$) ratio was constant for each object type respectively. $\texttt{inCont} = \{ \text{True}, \text{False} \}$ is not sampled but automatically determined after the stacking process.

\subsubsection{Sphere Dropping robot experiment setup}
We run and assess the sphere dropping experiment on the TIAGo service robot (as introduced in Sec.~\ref{sec:cubeStackingRobotExperiment}). As a sphere, we use a regular tennis ball, which weighs around 58g and has a diameter of around 6.6cm. We chose three containers which each possess 
a height and size parametrization that was captured in the model (Fig.~\ref{fig:environment2}.d). We evaluate each container at nine different points, where $\texttt{xOff}, \texttt{yOff} = \{0, 3, 6\}$ (in cm). For each unique sphere dropping setup instantiation, we conduct five iterations. Similar to the prior experiment, the containers are re-adjusted by a human operator, who is also determining the dropping outcome.


\section{Results and Discussion}
\label{sec:results}

\subsection{Analysis of the obtained causal models}
\begin{figure}[ht!]
\centering
  \includegraphics[width=0.4\textwidth]{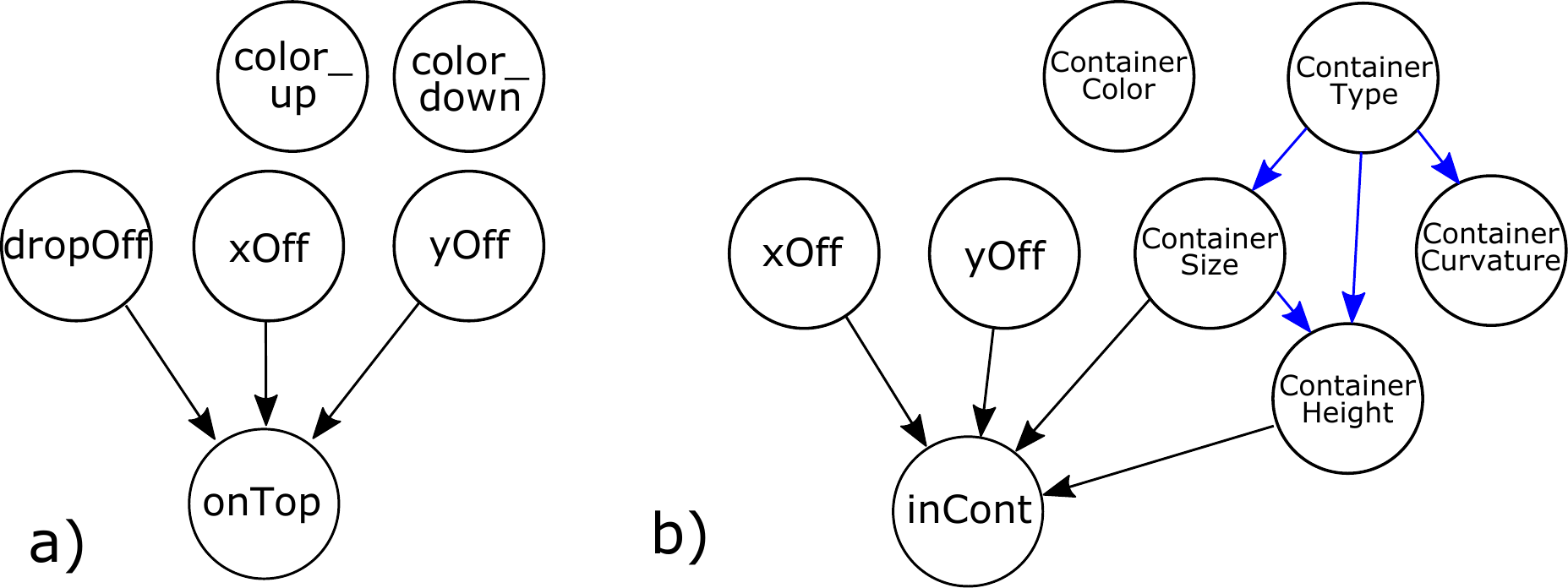}
  \caption{Obtained bayesian network structure for a) the cube stacking experiment and b) the sphere dropping experiment. The blue edges have been detected by the structure learning algorithm but had to be directed manually.}
\label{fig:bnGraph}
\end{figure}

We first present and discuss the learned causal model of the cube stacking scenario. 10-fold cross-validation reports an average loss of 0.10269 and a standard deviation of 0.00031, and Figure \ref{fig:bnGraph}.a displays the resulting DAG.
The graph indicates, that there are causal relations from \texttt{xOff}, \texttt{yOff} and \texttt{dropOff} to \texttt{onTop}, while the two color variables \texttt{colorDown} and \texttt{colorUp} are independent. In other words, it makes a difference from which position the cube is dropped, but the cube color has no impact on the stacking success.
We obtained the following \texttt{dropOff} intervals (in cm):
\begin{itemize}[leftmargin=0.5cm]
    \item[] {\footnotesize $z_1$:[$0.4,1.8$], $z_2$:($1.8,3.2$], $z_3$:($3.2,4.5$], $z_4$:($4.5,5.9$], $z_5$:($5.9,7.3$], $z_6$:($7.3,8.6$], $z_7$:($8.6,9.9$],}
\end{itemize}
and the following \texttt{xOff/yOff} intervals (in cm):
\begin{itemize}[leftmargin=0.5cm]
    \item[] {\footnotesize$x/y_1$:[$-3,-1.8$], $x/y_2$:($-1.8,-0.6$], $x/y_3$:($-.6,.6$], $x/y_4$:($.6,1.8$], $x/y_5$:($1.8,3$]}
\end{itemize}

\begin{figure}[ht!]
\centering
  \includegraphics[width=0.48\textwidth]{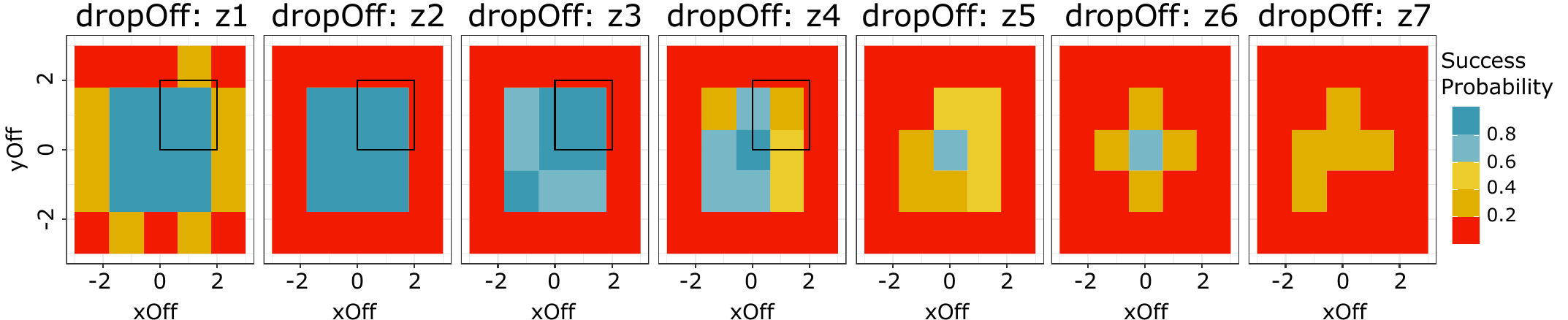}
  \caption{Visualisation of the conditional probability table for $P(\texttt{onTop}=1|\Pi_{\texttt{onTop}})$. \texttt{xOff}, \texttt{yOff} are discretized into 5 intervals and \texttt{dropOff} into 7. 
  Values are in cm. The black rectangles denote the \texttt{xOff} and \texttt{yOff} value range evaluated in the real-world experiments.}
\label{fig:stackSucc}
\end{figure}

The conditional probabilities $P(\texttt{onTop}=1|\Pi_{\texttt{onTop}})$ are visualized in Fig. \ref{fig:stackSucc}. These plots allow us to conclude that stacking success decreases the greater the drop-offset and the more offset in both x- and y-direction. In particular, there is a diminishing chance of stacking success for the values $|\text{xOff}|> 1.8$ or $|\text{yOff}|> 1.8$, no matter the dropOff height.

The obtained DAG for the sphere dropping experiment is visualized in Fig.~\ref{fig:bnGraph}.b). The algorithm detected causal links between $\texttt{contHeight}$ $\texttt{contSize}$, $\texttt{contType}$,  $\texttt{contCurvature}$ (marked in blue), but, initially, was not able to direct these four edges (which is a common problem in structure learning as discussed in Sec.~\ref{sec:structureLearning}). Since it is not possible to fit a conditional probability table based on an undirected graph, we directed these edges manually based on our domain knowledge of the task. 10-fold cross validation reports an average loss of 0.184 and a standard deviation of 0.000052. The graph indicates causal links between $\texttt{contType}$ and $\texttt{inCont}$ via $\texttt{contHeight}$ and $\texttt{contSize}$, while $\texttt{contColor}$ and $\texttt{contCurvature}$ do not impact the dropping success.


The following table shows the obtained intervals for the four variables that affect \texttt{inCont} (all values are in cm): 
\begin{table}[h]
\centering
\begin{tabular}{|l|l|l|}
\hline
\texttt{ContSize} & \texttt{ContHeight} & \texttt{x/yOff}  \\
\hline
\hline
$s_1:[6.3,11.2]$ & $h_1:[1.5,2.9]$ & $x/y_1:[-6,-3.6]$  \\
\hline
$s_2:(11.2,14]$ & $h_2:(2.9,6.3]$ & $x/y_2:(-3.6,-1.2]$  \\
\hline
$s_3:(14,17]$ & $h_3:(6.3,9.5]$ & $x/y_3:(-1.2,1.2]$  \\
\hline
$s_4:(17,22]$ & $h_4:(9.5,12.3]$ & $x/y_4:(1.2,3.6]$  \\
\hline
$s_5:(22,30]$ & $h_5:(12.3,19]$ & $x/y_5:(3.6,6]$   \\
\hline
\end{tabular}
\end{table}

\begin{figure}[ht!]
\centering
  \includegraphics[width=0.48\textwidth]{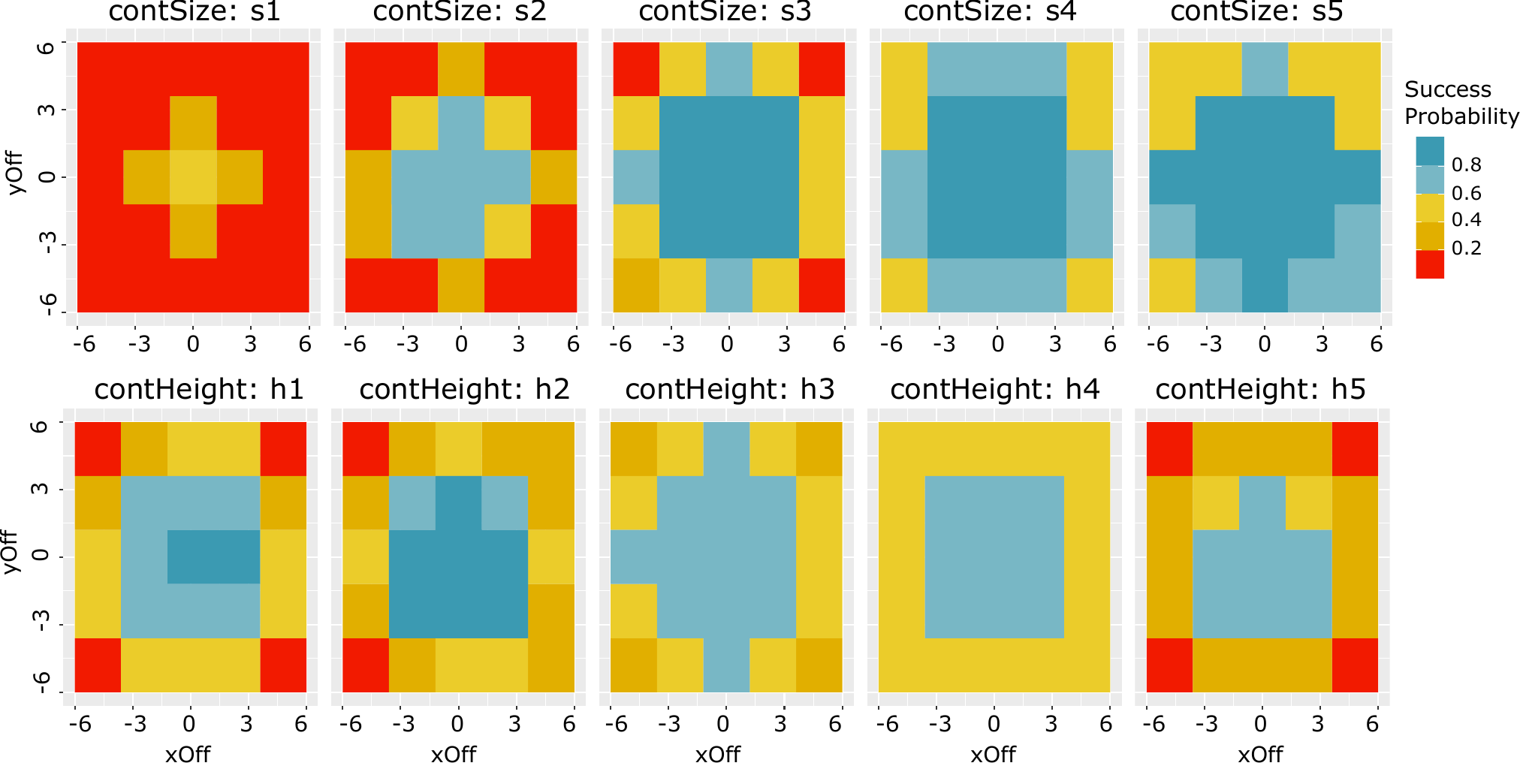}
  \caption{Visualisation of the conditional probability table for $P(\texttt{inside}=1|\Pi_{\texttt{inside}})$. All values are in cm.}
\label{fig:stackSucc2}
\vspace{-5mm}
\end{figure}

Querying the causal model for sphere dropping success given the object type reveals that the task is most likely successful for bowls (69\%) followed by plates (59\%) and glasses (25\%). The model also indicates that larger-sized objects are more tolerant to x/y-offsets and yield a higher chance of dropping success. 
(Fig.~\ref{fig:stackSucc2}). Surprisingly, a similar trend cannot be determined with the container height (Fig.~\ref{fig:stackSucc2}). The reasons are twofold: First, the container height depends not only on the size but also on the type of container (e.g., glasses typically have a larger height than plates). As a result, not all object types are represented in all height intervals, e.g., plates are only covered in h1 and h2, whereas cups are distributed among h2-h4. Second, cups and plates were found to contribute to a higher dropping success chance. As a result, the largest success chances can be obtained in interval h2.

Overall, we conclude that the obtained success probabilities resemble our intuitive understanding of the physical processes for both scenarios. Nevertheless, real-world experiments have a higher complexity due to the many environment uncertainties. We, therefore, expect the simulation to be less conservative than reality, as we have higher control over the variables involved in the stacking process.

\subsection{sim2real accuracy of the causal models}
To evaluate how well the causal model and the real-world match, we introduce 
the sim2real accuracy score. It is defined as the normalized difference in predicted probabilities over the set of points that were evaluated in real-world experiments. 

\begin{figure}[ht!]
\centering
  \includegraphics[width=.44\textwidth]{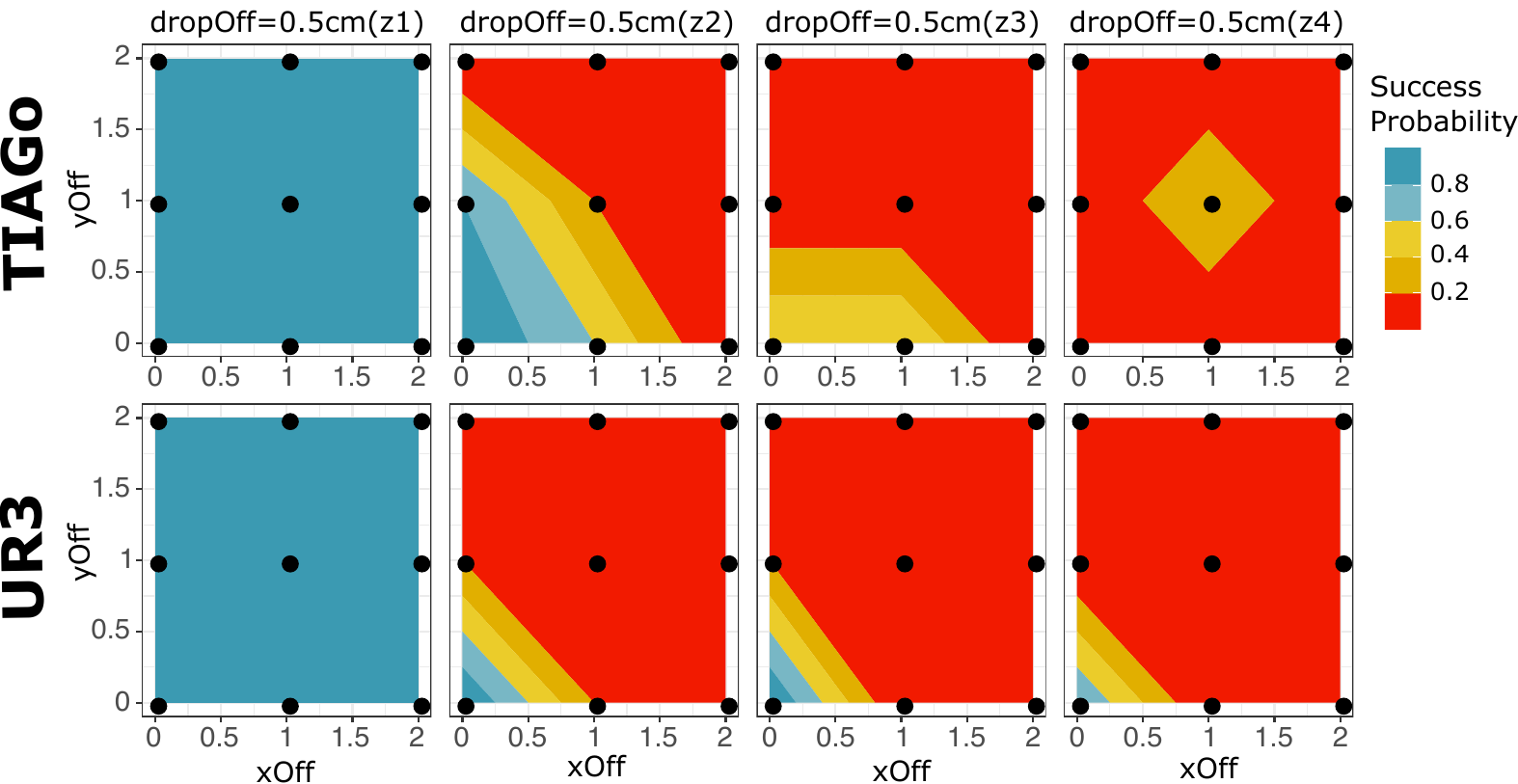}
  \caption{Shows real-world success probabilities for the task of stacking cubes, evaluated for two robots (TIAGo and UR3) at 36 different points (black dots). The probabilities are interpolated between the nine measurement points for each $\texttt{dropOff}$ value. All values are in cm.}
\label{fig:experimentComparison}
\vspace{-5mm}
\end{figure}

The results for the real-world experiments of the cube stacking scenario are presented in Fig. \ref{fig:experimentComparison}, where the black points indicate the nine stacking locations (all possible combinations of x- and y-offset values) for each of the four drop-off heights. The plots show the contours of the probabilities, meaning the stacking success probabilities are interpolated between the nine measurement points. 
The sim2real accuracy amounts to 71\% for the TIAGo and 69\% for the UR3. The largest discrepancy between model and reality can be determined for the higher drop-off positions. For the real-world measurements, the stacking success drops earlier, at around 2cm or 3.5cm.
It is also interesting to compare similarities regarding probability outcomes between the two differently embodied robots. The correspondence concerning the 36 measured positions amounts to 85\%. 

\begin{figure}[ht!]
\centering
  \includegraphics[width=0.43\textwidth]{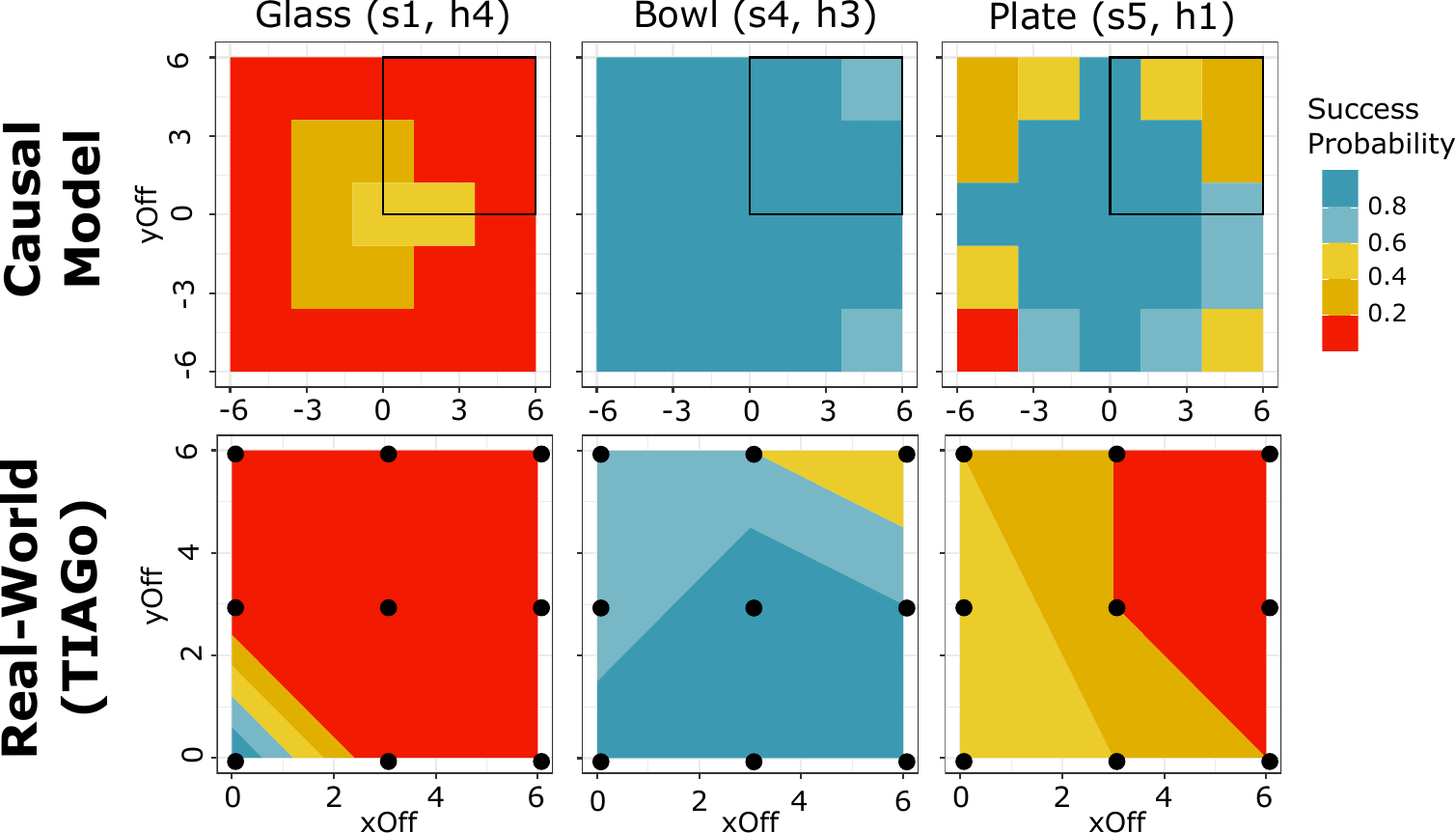}
  \caption{Compares success probabilities for the task of sphere dropping between the real experiments (evaluated at 9 different points marked as black dots) and the causal model. The black rectangles (causal model) denote the \texttt{xOff} and \texttt{yOff} value range (in cm) evaluated in the real-world experiments.}
\label{fig:experimentComparison2}
\vspace{-5mm}
\end{figure}

The results for the sphere dropping experiments are presented in Fig. \ref{fig:experimentComparison2}. We obtained 72\% sim2real accuracy for the tested data points. We observed the most significant discrepancy for the plate, which was predicted to have a much larger success probability by the model. One reason could have been, for example, that the surface of the real plate was not perfectly flat as in the simulations. 

We can conclude that for both tested scenarios, the probability model obtained from simulated data matches reasonably well with reality and thus can be utilized for the explanation of failures that occur in the real world. Furthermore, the model generalizes well to differently embodied robots. We want to emphasize that the causal model was not retrained or adapted when the real scenarios were tested. 
If we had obtained a lower sim2real accuracy or more significant differences between the two robots, it would be advisable to include robot-specific variables (such as the gripper type and orientation) and adapt the model with real-world data. But even then, the model that we obtain from the simulation can be used as an excellent experience prior, allowing for faster applicability and learning. 



\subsection{Explainability capabilities}
\begin{figure}[ht!]
\vspace{-3mm}
\centering
  \includegraphics[width=0.35\textwidth]{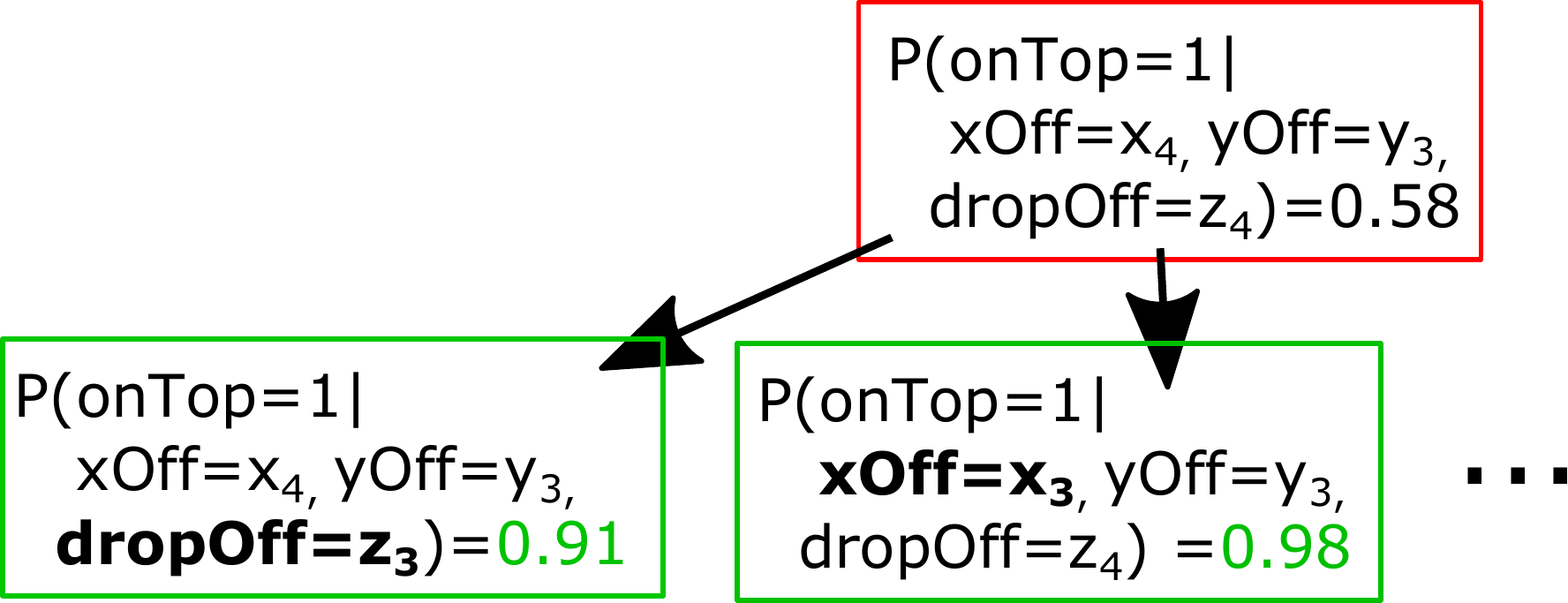}
  \caption{BFS for explaining Cube Stacking - Example 2 from Tab.~\ref{tab:examples}}
\label{fig:bfs}
\vspace{-3mm}
\end{figure}

Finally, we provide several concrete examples to showcase which kind of explanations our method finds for robot failures. We set the probability threshold that distinguishes a failure from success to $\epsilon = 0.8$ for all examples. Tab.~\ref{tab:examples} provides three examples for both scenarios of stacking a cube and dropping a sphere. Cube Stacking - Example 2 is particularly interesting, as it showcases that there are often multiple correct explanations for the error. In this case it would have been possible to achieve a successful stacking by either going from $\texttt{dropOff}=z_4$ to $\texttt{dropOff}=z_3$ or by changing $\texttt{xOff}=z_4$ to $\texttt{xOff}=z_3$ (search tree is visualised in Fig. \ref{fig:bfs}). Which solution is found first depends on the variable prioritization within the tree search due to the used BFS algorithm. 

\begin{table}[h]
\begin{tabular}{|l|l|l|l|l|}
\hline
input & \begin{tabular}[c]{@{}l@{}}input\\interval\end{tabular} & \begin{tabular}[c]{@{}l@{}}curr. succ. \\ probability \end{tabular} & \begin{tabular}[c]{@{}l@{}} closest \\solution\\ interval \end{tabular}& \begin{tabular}[c]{@{}l@{}}exp. succ. \\ probability \end{tabular}\\
\hline
\hline
\multicolumn{5}{|c|}{\textbf{Cube Stacking - Example 1:}}\\
\hline
\begin{tabular}[c]{@{}l@{}}xOff = 0.0\\ yOff = 0.02\\ dropOff = 0.08\end{tabular} & \begin{tabular}[c]{@{}l@{}}$x_3$\\ $\boldsymbol{y_5}$\\ $\boldsymbol{z_6}$\end{tabular}  & 0.0 & \begin{tabular}[c]{@{}l@{}}$x_3$\\ $\boldsymbol{y_4}$\\ $\boldsymbol{z_3}$\end{tabular} & 0.85 \\
\hline
\multicolumn{5}{|l|}{\begin{tabular}[c]{@{}l@{}}\textbf{Explanation}: The upper cube was stacked too high and too far \\ to the front of the lower cube.\end{tabular}} \\
\hline
\hline
\multicolumn{5}{|c|}{\textbf{Cube Stacking - Example 2:} }\\
\hline
\begin{tabular}[c]{@{}l@{}}xOff = 0.015\\ yOff = 0.0\\ dropOff = 0.05\end{tabular} & \begin{tabular}[c]{@{}l@{}}$x_4$\\ $y_3$\\ $\boldsymbol{z_4}$ \end{tabular}  & 0.58 & \begin{tabular}[c]{@{}l@{}}$x_4$\\ $y_3$\\ $\boldsymbol{z_3}$ \end{tabular} & 0.91 \\
\hline
\multicolumn{5}{|l|}{\textbf{Explanation}: The upper cube was stacked too high.} \\
\hline
\hline
\multicolumn{5}{|c|}{\textbf{Cube Stacking - Example 3: }}\\
\hline
\begin{tabular}[c]{@{}l@{}}xOff = -0.015\\ yOff = 0.015\\ dropOff = 0.02\end{tabular} & \begin{tabular}[c]{@{}l@{}}$\boldsymbol{x_1}$\\ $\boldsymbol{y_1}$\\ $z_2$\end{tabular}  & 0.017 & \begin{tabular}[c]{@{}l@{}}$\boldsymbol{x_2}$\\ $\boldsymbol{y_2}$\\ $z_2$\end{tabular} & 1.0 \\
\hline
\multicolumn{5}{|l|}{\begin{tabular}[c]{@{}l@{}}\textbf{Explanation}: The upper cube was stacked too far to the left and \\  too far to the back of the lower cube.\end{tabular}} \\
\hline
\hline
\multicolumn{5}{|c|}{\textbf{Sphere Dropping - Example 1:}}\\
\hline
\begin{tabular}[c]{@{}l@{}}xOff = 0.015\\ yOff = 0.015\\ ContSize = .08 \\ ContHeight = .1\end{tabular} & \begin{tabular}[c]{@{}l@{}}$x_4$\\ $y_4$\\ $\boldsymbol{s_1}$ \\ $\boldsymbol{h_4}$\end{tabular}  & 0.189 & \begin{tabular}[c]{@{}l@{}}$x_4$\\ $y_4$ \\ $\boldsymbol{s_3}$ \\ $\boldsymbol{h_3}$ \end{tabular} &  0.957 \\
\hline
\multicolumn{5}{|l|}{\begin{tabular}[c]{@{}l@{}}\textbf{Explanation}: The container was too small and too high (interpretation:\\take a bowl instead of a cup, which is bigger but has less height)\end{tabular}} \\
\hline
\hline
\multicolumn{5}{|c|}{\textbf{Sphere Dropping - Example 2:} }\\
\hline
\begin{tabular}[c]{@{}l@{}}xOff = 0.059 \\ yOff = 0.059 \\ ContSize = .21 \\ ContHeight = .09 \end{tabular} & \begin{tabular}[c]{@{}l@{}}$\boldsymbol{x_5}$\\ $y_5$ \\ $s_4$  \\ $h_3$ \end{tabular}  & 0.727 & \begin{tabular}[c]{@{}l@{}}$\boldsymbol{x_4}$\\ $y_5$ \\ $s_4$  \\ $h_3$ \end{tabular} & 0.98 \\
\hline
\multicolumn{5}{|l|}{\textbf{Explanation}: The sphere was dropped too far to the right.} \\
\hline
\hline
\multicolumn{5}{|c|}{\textbf{Sphere Dropping - Example 3: }}\\
\hline
\begin{tabular}[c]{@{}l@{}}xOff = -0.03 \\ yOff = -0.03 \\ ContSize = .15 \\ ContHeight = .017 \end{tabular} & \begin{tabular}[c]{@{}l@{}}$x_2$\\ $y_2$ \\ $\boldsymbol{s_3}$ \\ $h_1$\end{tabular}  & 0.58 & \begin{tabular}[c]{@{}l@{}}$x_2$\\ $y_2$ \\ $\boldsymbol{s_4}$ \\ $h_1$\end{tabular} & 0.933 \\
\hline
\multicolumn{5}{|l|}{\begin{tabular}[c]{@{}l@{}}\textbf{Explanation}: The container (plate) was too small. \end{tabular}} \\
\hline
\end{tabular}
\caption{Three examples of failure explanations for both scenarios of Cube Stacking and Sphere Dropping. Intervals that were subject to changes in the closest solution (and thus contributed to the failure explanation) are marked in bold letters.}
\label{tab:examples}
\end{table}

Our closest solution will lead to a minimal number of interval changes and thus provides the 'simplest' solution in terms of Occam's razor principle~\cite{Miller19Explanation}. For instance, in Sphere Dropping - Example 2, it would have also been possible to change the container to a larger bowl. But instead, the search process found it was easier to adapt the \texttt{xOff} position. The advantage of the current uninformed BFS is that this principle is always applicable and does not require any human domain knowledge. 

\subsection{Comparison of our failure explanation approach with baseline methods}
\label{sec:baseline}
\begin{table*}[]
\begin{tabular}{|l|l|l|l|l|l|}
\hline
 & 
\begin{tabular}[c]{@{}l@{}}Method\end{tabular} & 
Output & 
Detection of caus. relevant variables & 
\begin{tabular}[c]{@{}l@{}}Learning  
Prerequisites\end{tabular} & 
Task Succ. Predict.   \\
\hline
\hline
\begin{tabular}[c]{@{}l@{}}CB-H\\\cite{Das21}\end{tabular}& 
\begin{tabular}[c]{@{}l@{}} Fault trees + encoder-\\decoder network\end{tabular} & 
\begin{tabular}[c]{@{}l@{}}language model/ spoken \\ failure explanation\end{tabular} & 
\begin{tabular}[c]{@{}l@{}}no differentiation between causally \\relevant and irrelevant env. variables\end{tabular} & 
\begin{tabular}[c]{@{}l@{}}  failure-cause annotated \\ simulations\end{tabular} & 
\begin{tabular}[c]{@{}l@{}}no\end{tabular}                                                              \\
\hline
\begin{tabular}[c]{@{}l@{}}SSG-R\\\cite{Das21b}\end{tabular}& 
\begin{tabular}[c]{@{}l@{}} MOTIFNET~\cite{Zellers18}\\+ pairwise ranking\end{tabular} & 
\begin{tabular}[c]{@{}l@{}}list of relevant \\ spatial and object relations\end{tabular} & 
\begin{tabular}[c]{@{}l@{}}informally, through pairwise ranking\end{tabular} & 
\begin{tabular}[c]{@{}l@{}}relationship ranking \\labels\end{tabular} & 
\begin{tabular}[c]{@{}l@{}}no\end{tabular} \\
\hline
ours & 
\begin{tabular}[c]{@{}l@{}}causal BNs +\\contrastive BFS \end{tabular}& 
\begin{tabular}[c]{@{}l@{}}contrastive failure \\variable parametrizations \end{tabular}& 
\begin{tabular}[c]{@{}l@{}}formally, through BN \\structure learning \end{tabular}& 
\begin{tabular}[c]{@{}l@{}}samples $d$ (including \\action
outcome)\end{tabular}& 
\begin{tabular}[c]{@{}l@{}}MLE (or similar \\ like Bayesian est.)\end{tabular} \\
\hline
\end{tabular}
\caption{Comparison of our explanation generation pipeline with other approaches.}
\label{tab:baseline}
\vspace{-5mm}
\end{table*}

We compare our method of finding explanations of robot task failures with the two closely related methods of Context-Based History (CB-H)~\cite{Das21} explanations, and the ranked Semantic Scene Graph method (SSG-R)~\cite{Das21b}, based on the criteria that are summarized in Tab.~\ref{tab:baseline}. For CB-H all failures and their causes need to be manually defined in the form of Fault Trees. In SSG-R failures are not modeled, but explained in form of a list of spatial relations (like \textit{close to} or \textit{occluded}) and object features (like \textit{fragile} or \textit{heavy}), automatically detected through the semantic scene graph model MOTIFNET~\cite{Zellers18}. We explain failures via contrastive variable parametrizations. Due to these differences in failure representation, all three methods have different requirements during the learning step. For learning the encoder-decoder network that generates language failure explanations for CB-H, simulations must be annotated with the respective failure cause. In~\cite{Das21}, 2100 annotated time-steps were used to train for six different failure causes. However, the number of required samples will drastically increase for the two discussed examples of cube stacking and sphere dropping due to the increased number of failure possibilities. Additionally, samples are more expensive than in our method since it is required to label the failure cause instead of a simple binary action success label. In SSG-R, pairwise ranking distinguishes between relevant and irrelevant relations.
Pairwise relation preferences must be provided via domain knowledge of the failure scenario and which are more expensive than the automatically retrievable binary action success labels from our method. Another difficulty in terms of applicability to the presented scenarios of cube stacking and sphere dropping provide the continuous variables (e.g., \texttt{contSize} or \texttt{xOff}), which are discretized into more than two categories (as opposed to binary object relations). For these variables, MOTFNET is not applicable. While, in principle, a range of variables was detected to influence the action outcome causally, it is due to a specific variable parametrization that they lead to the action failure. Our method automatically discerns between relevant and irrelevant relations. Last but not least, neither CB-H nor SSG-R learn an action success model, which can be useful for other tasks beyond failure explanation, e.g., failure prediction and prevention.

To conclude, both of these methods would require significant changes and adaption to find explanations for the experiment scenarios discussed in this paper. One of the most significant differences of both methods with ours is the requirement of failure-cause labels instead of action success labels, which are typically easier to obtain.

\section{Conclusion}
This paper presents our novel approach to finding causal explanations for robot failures. First, we learn a causal Bayesian network from simulated task executions. We show that the model is transferable to the real world with 70\% and 72\% accuracy over two tasks of stacking cubes and dropping a sphere into different containers and is agnostic to differently embodied robots. Furthermore, we propose a new method to generate explanations of execution failures based on the causal model. This method finds a contrastive explanation comparing the action parametrization of the failure with its closest parametrization that would have led to a successful execution, which is found through breadth-first search (BFS). For future work, we would like to incorporate a language model that automatically encodes the contrastive failure explanations into a vector of words, such that it can be communicated more intuitively to a wide range of potential users. Furthermore, we want to investigate how the obtained causal models can also be used to predict and prevent failures from happening.

\addtolength{\textheight}{-12cm}   




\section*{Acknowledgment}
\addcontentsline{toc}{section}{Acknowledgment}
The research reported in this paper has been supported by Chalmers AI Research Centre (CHAIR).

\bibliography{mybib}
\bibliographystyle{IEEEtran}

\end{document}